%% file: main.tex
\documentclass[10pt,twocolumn,letterpaper]{article}

\usepackage[pagenumbers]{cvpr}
\usepackage{lipsum}

\input{preamble}

\definecolor{cvprblue}{rgb}{0.21,0.49,0.74}
\usepackage[pagebackref,breaklinks,colorlinks,allcolors=cvprblue]{hyperref}

\def\confName{CVPR}
\def\confYear{2026}

\title{Recovering Physically Plausible Human-Object Interactions \\ from Monocular Videos}

\author{
    Dingbang Huang$^{1,2}$
    \quad
    Etienne Vouga$^{1}$
    \quad
    Qixing Huang$^{1}$
    \quad
    Georgios Pavlakos$^{1}$
    \\[1.5mm]
    $^{1}$University of Texas at Austin
    \quad
    $^{2}$Shanghai Jiao Tong University
}

\begin{document}

\twocolumn[{%
\renewcommand\twocolumn[1][]{#1}%
\maketitle
\begin{center}
    \newcommand{\teaserwidth}{\textwidth}
    \vspace{-0.8em}
    \centerline{
   \includegraphics[width=\teaserwidth,clip]{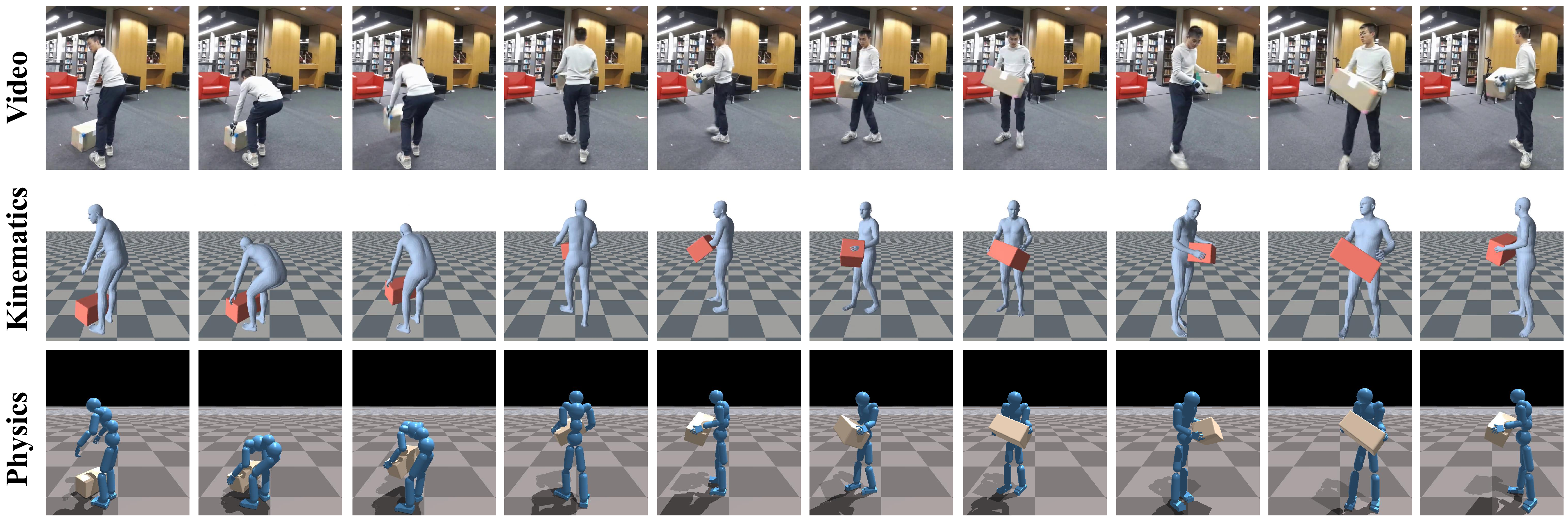}
     }
\vspace{-0.5em}
   \captionof{figure}{\textbf{Physically plausible reconstruction of human-object interactions from monocular video.} Given an input video, we start from a noisy kinematic reconstruction (e.g., incorrect contact, floating objects, etc).
   Then, we optimize a policy for this sequence that can rollout a physically plausible version of the observed interaction.}
   \vspace{-0.8em}
\label{fig:teaser}
\end{center}%
}]

\input{sec/0_abstract}
\input{sec/1_introduction}
\input{sec/2_related}
\input{sec/3_method}
\input{sec/4_experiments}
\input{sec/5_conclusion}

{
    \small
    \bibliographystyle{ieeenat_fullname}
    \bibliography{main}
}

\end{document}

%% file: preamble.tex
\usepackage{pifont}
\usepackage{colortbl}
\usepackage{makecell}
\usepackage{multirow}

\definecolor{darkgreen}{HTML}{2E7D32}
\definecolor{lightgreen}{HTML}{E8F5E9}
\definecolor{darkred}{HTML}{C62828}
\definecolor{lightred}{HTML}{FFEBEE}
\definecolor{darkorange}{HTML}{E65100}
\definecolor{lightorange}{HTML}{FFF3E0}

\newcommand{\cmark}{{\color{darkgreen}\ding{51}}}
\newcommand{\xmark}{{\color{darkred}\ding{55}}}
\newcommand{\yes}{\cellcolor{lightgreen}\cmark}
\newcommand{\no}{\cellcolor{lightred}\xmark}
\newcommand{\pmark}{\cellcolor{lightorange}{\color{darkorange}(\ding{51})}}

%% file: sec/0_abstract.tex
\begin{abstract}
In this paper, we propose \textbf{RePHO}, a method to reconstruct physically plausible human-object interactions (HOI) from monocular videos.
While existing kinematic-based approaches produce visually plausible motion, they often result in physically implausible artifacts such as interpenetration and object floating.
To overcome these issues, we introduce a physics-guided reconstruction framework.
We begin with a kinematic estimate and then refine it by training a policy with reinforcement learning (RL).
This policy is optimized to reproduce the interaction in a physics simulator.
Because kinematic estimates are typically noisy, naive RL training can fail. Therefore, we propose an adaptive sampling strategy with a dual self-updating mechanism that can identify the frames with the most informative and reliable kinematic reconstruction.
Our process progressively improves reconstruction quality and yields physically consistent HOI sequences.
We demonstrate our approach on two standard HOI benchmarks and achieve clear improvements in physical plausibility metrics over state-of-the-art methods. Project Page:\href{https://dingbang777.github.io/RePHO/}{https://dingbang777.github.io/RePHO/}
\end{abstract}

%% file: sec/1_introduction.tex
\section{Introduction}
\label{sec:introductin}
Reconstructing whole-body human-object interactions (HOI) from images and videos is a long-standing challenge in 3D computer vision.
Recent progress in monocular HOI reconstruction has enabled impressive visual results from RGB videos~\cite{xie2023visibility,xie2024template,xie2025intertrack}.
However, these methods remain kinematic in nature and often exhibit physical implausibilities, such as floating contacts, interpenetration, and jittering.
While some approaches introduce physics-aware losses to penalize contact violations, they still fall short of true physically plausible results, as they do not explicitly model contact forces, gravity and collisions that govern real-world interactions.

To address this limitation, we propose RePHO (\textbf{Re}constructing \textbf{P}hysically Plausible \textbf{H}uman-\textbf{O}bject Interactions), a simulation-based framework that refines noisy kinematic reconstructions using reinforcement learning (RL) within a physics simulator.
Our idea is intuitive: we start from a kinematic reconstruction and then train a control policy to reproduce the motion under physical laws.
However, a key challenge lies in the extreme noise present in monocular reconstructions, especially during occlusions or fast motions, where estimated object poses may drift or become unstable.
As a result, naively training an RL policy on such noisy sequences often leads to unstable or failed behaviors.
Existing RL-based HOI frameworks~\cite{wang2023physhoi,wang2025skillmimic,xu2025intermimic} assume access to clean motion capture data, which lacks these artifacts and thus cannot directly operate on monocular reconstructions.

Despite this noise, monocular reconstructions still contain useful signals. For instance, frames with visible contacts or slow object motion tend to be more accurate.
Our key insight is to identify and exploit these reliable segments to guide learning.
To that end, we introduce a novel adaptive sampling and dual self-updating mechanism that automatically selects informative frames as anchors and progressively propagates physically consistent motion across the entire sequence.
This adaptive strategy enables the policy to learn stable and realistic interactions, even when the input reconstructions are severely degraded.

Our contributions can be summarized as follows:

\begin{itemize}
\item We propose RePHO, a simulation-based framework for reconstructing physically plausible human-object interactions directly from RGB input.

\item To handle the severe noise of kinematic estimates, we introduce a novel strategy that captures informative motion cues and iteratively refines both kinematic and physical states for improved consistency.

\item Our method significantly improves physical plausibility metrics across two standard HOI benchmarks compared to state-of-the-art monocular approaches.
\end{itemize}

%% file: sec/2_related.tex
\section{Related Work}
\label{sec:related}

\subsection{3D Human-Object Interaction Reconstruction}
Previous work on the perception of 3D human-object interactions was largely driven by the development of datasets capturing human-object kinematics.
For example, BEHAVE~\cite{bhatnagar2022behave} and InterCap~\cite{huang2022intercap} provide RGB-D or multi-view sequences with accurate human-object geometry and contact annotations, supporting supervised learning and evaluation.
More recent datasets~\cite{zhao2024m,zhang2023neuraldome,liu2022hoi4d,banerjee2025hot3d,wen2025reconstructing,jiang2023full,taheri2020grab,lu2025humoto,kim2025parahome} greatly expand object and scene diversity, temporal coverage, and motion complexity, enabling reconstruction frameworks to address in-the-wild and multi-person scenarios.

Building on these datasets, image-based HOI reconstruction methods aim to infer 3D human and object geometry, as well as their spatial relations, from a single image.
PHOSA~\cite{zhang2020perceiving} jointly optimizes human and object poses under reprojection and contact constraints, while subsequent works~\cite{xie2022chore,huo2024monocular,cseke2025pico,wen2025reconstructing,dwivedi2025interactvlm} extend this paradigm
with learned contact priors and occlusion-aware reasoning.

In contrast, video-based methods leverage temporal cues for
consistent 4D reconstruction.
VisTracker~\cite{xie2023visibility} introduces a template-based framework that tracks human and object meshes across frames via interaction-conditioned neural fields.
Follow-up works~\cite{xie2024template,xie2025intertrack} improve temporal coherence by combining motion priors and cross-frame correspondences, enabling robust tracking under occlusions and appearance changes without relying on object templates.
More recently, zero-shot frameworks~\cite{li2024zerohsi,li2025hoi,kim2025david,lou2025zero,xie2026cari4d,wen2025efficient,li2026anylift} have coupled video-diffusion motion priors with contact-aware optimization, enabling reconstruction across unseen object categories and interaction types.

\subsection{Physics-Based Interaction Animation and Motion Imitation}
Early approaches on physics-based human motion imitation relied on state machines or handcrafted control objectives, which restricted generalization beyond specific motion tasks~\cite{hodgins1995animating,yin2007simbicon,da2008interactive}.
With the advent of reinforcement learning (RL) and large-scale motion datasets, locomotion imitation has advanced dramatically, integrating adversarial priors, hierarchical controllers, and generative conditioning to reproduce diverse and highly dynamic whole-body behaviors~\cite{peng2018deepmimic,peng2021amp,peng2022ase,luo2022embodied,luo2023perpetual,truong2024pdp,liao2025beyondmimic,huang2025diffuse,wu2025uniphys,luo2025sonic,ugrinovic2024multiphys}.

Extending these successes to human-object interactions introduces a much higher level of complexity.
Unlike pure locomotion, HOI requires modeling multi-contact events, tight body-object coupling, and robustness to imperfect motion data.
To address these challenges, recent works in physics-based HOI imitation emphasize unified reward formulations, contact-aware representations, and data-driven generalization mechanisms~\cite{wang2023physhoi,wang2025skillmimic,xu2025intermimic,braun2024physically}.
Rather than handcrafting task-specific reward terms~\cite{hassan2023synthesizing,yuan2023learning,bae2023pmp}, modern frameworks encode interaction dynamics through generalizable physical signals, such as relative body-object motion and contact structure.
This paradigm enables policies to reproduce a wide range of HOI behaviors, from dynamic manipulation to full-body loco-manipulation, within a unified control framework~\cite{gao2024coohoi,tessler2024maskedmimic,pan2025tokenhsi}.
Meanwhile, scalable training pipelines increasingly integrate imitation learning and teacher-student distillation, converting large collections of motion capture (MoCap) sequences into physically plausible demonstrations for general-purpose interaction policies.
Together, these developments reflect a broader trend toward unifying motion tracking, contact reasoning, and control within a single physics-consistent learning framework, pushing HOI animation closer to simulation-ready universal interaction models~\cite{xu2025intermimic}.

Despite this progress, applying physics-based motion imitation to in-the-wild monocular reconstructions remains
challenging.
While prior tracking approaches have successfully retargeted human-only motions from videos~\cite{peng2018deepmimic,luo2023perpetual,ugrinovic2024multiphys,feng2025physhmr}, extending these methods to HOI is far more demanding: HOI tasks impose stricter accuracy requirements, yet the visual reconstructions they rely on are typically noisy and incomplete.

%% file: sec/3_method.tex
\section{Method}
\label{sec:method}

\begin{figure*}
\centering
\vspace{-0.5em}
\includegraphics[width=1.0\textwidth]{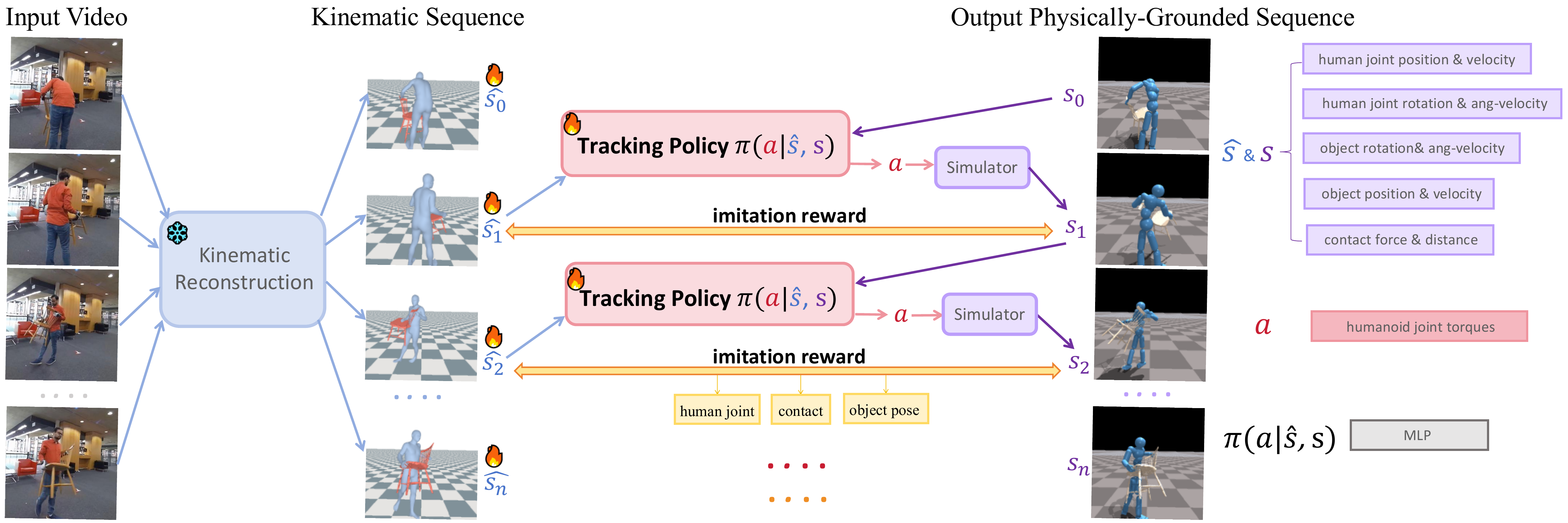}
\vspace{-1.5em}
\caption{
\textbf{Overview of our two-stage pipeline.}
In the first stage, we use off-the-shelf kinematic reconstruction method VisTracker~\cite{xie2023visibility} to reconstruct human-object interactions in global coordinates from the input video, producing kinematic estimates that are often noisy.
In the second stage, we train a \emph{physics-based tracking policy} to imitate these reference kinematics within a simulator using reinforcement learning.
At each timestep $t$, the policy receives the current physical state $s_t^s$ and a set of future reference states $\{ \hat{s}_{t,t+k} \}_{k \in \mathbf{K}}$, and outputs an action $a_t$ that drives the humanoid toward the next kinematic target while maintaining physical plausibility.
Rolling out this process over time yields a \emph{physically grounded} human-object interaction sequence.
}
\vspace{-1.5em}
\label{fig:HOI_tracking}
\end{figure*}

\subsection{Problem Formulation}
Given a monocular video $I_{1:T}$ consisting of $T$ frames that capture a person interacting with an object, our goal is to recover the human and object motions
in a \textit{physically plausible} manner. The reconstructed motion is denoted as
$M = \{ \mathbf{q}_{t}^{h}, \mathbf{q}_{t}^{o} \}_{t=1}^{T}$,
where $\mathbf{q}_{t}^{h}$ and $\mathbf{q}_{t}^{o}$ represent the estimated human and object states at frame $t$, respectively.

The human state $\mathbf{q}_{t}^{h}$ is parameterized using the SMPL-H model~\cite{MANO:SIGGRAPHASIA:2017}, which consists of the global orientation $\Phi_{t}^{h} \in \mathbb{R}^{3}$, body pose $\Theta_{t}^{h} \in \mathbb{R}^{J \times 3}$, body shape $\beta^{h} \in \mathbb{R}^{10}$, and root translation $\Gamma_{t}^{h} \in \mathbb{R}^{3}$. Each object is represented by its 6DoF pose
$\mathbf{q}_{t}^{o} = \{ \mathbf{R}_{t}^{o}, \mathbf{T}_{t}^{o} \}$,
where $\mathbf{R}_{t}^{o} \in \mathbb{R}^{3}$ and $\mathbf{T}_{t}^{o} \in \mathbb{R}^{3}$
denote its rotation and translation in the world coordinate system.

\subsection{Background on VisTracker}
VisTracker~\cite{xie2023visibility} is a state-of-the-art template-based method for reconstructing and tracking 3D human-object interactions.
Given an RGB video and corresponding human-object masks, it aims to recover temporally consistent human and object trajectories, along with their contact relationships.
The approach introduces the SMPL-T conditioned Interaction Field Network (SIF-Net), which learns neural fields conditioned on estimated SMPL meshes in camera space, enabling coherent 4D tracking through temporally consistent relative motion.
To handle occlusions, VisTracker also employs a Human-Object Visibility and Occlusion Prediction Network (HVOP-Net) that infers object poses using human motion cues and predicted visibility states, allowing robust tracking even when the object is partially or fully occluded.
Finally, the method jointly refines the human and object reconstructions under image-based and contact-based constraints, achieving high-quality recovery of complex human-object interactions.

\subsection{HOI Tracking with Reinforcement Learning}
\label{sec:hoi_tracking}
\noindent\textbf{Overview.}
We formulate HOI tracking as a Markov Decision Process (MDP) consisting of a state space, an action space, simulator-driven transition dynamics, and a reward function.
This formulation enables a learned policy to reproduce physically realistic human-object interactions through sequential decision making.

\medskip
\noindent\textbf{State.}
At each timestep $t$, the policy observes a state
\(
\mathbf{s}_t = \{ s_t^s, s_t^g \},
\)
which encodes the current physical state and a set of future kinematic reference states.
The physical state $s_t^s$ includes proprioceptive and object-related quantities:
\[
\big\{
\{ \theta_t^h, p_t^h, \dot{p}_t^h, \omega_t^h \},
\{ \theta_t^o, p_t^o, \dot{p}_t^o, \omega_t^o \},
(d_t, c_t)
\big\}.
\]
Here, $\theta$, $p$, $\dot{p}$, and $\omega$ denote joint rotations, positions, linear velocities, and angular velocities for the human ($h$) and object ($o$).
Following InterMimic~\cite{xu2025intermimic}, we also include two geometric/haptic cues: (i) $d_t$: vectors from human joints to nearest points on the object surface, and (ii) $c_t$: binary contact markers indicating whether a human body part exerts force on the object.
We provide contact markers only for the hands, as this minimal contact-guidance is necessary for stable HOI tracking even on clean MoCap data~\cite{wang2023physhoi,wang2025skillmimic}.
Contact markers on other human parts, contact points on the object and ground contact are not provided. Contact points on the object are instead acquired automatically through reinforcement learning and our proposed adaptive sampling and self-updating system.

The goal state
$s_t^g = \{ \hat{s}_{t,t+k} \}_{k \in \mathbf{K}}$
aggregates future kinematic references. Each element $\hat{s}_{t,t+k}$
encodes the reference at time $t{+}k$ relative to the current physical state at $t$:
\begin{equation}
\begin{aligned}
\big\{
& \{ \hat{\theta}_{t+k}^h \ominus \theta_t^h,\ \hat{p}_{t+k}^h - p_t^h \},~
  \{ \hat{\theta}_{t+k}^o \ominus \theta_t^o,\ \hat{p}_{t+k}^o - p_t^o \}, \\
& \{ \hat{d}_{t+k} - d_t,\ \hat{c}_{t+k} - c_t \},~
  \{ \hat{\theta}_{t+k}^h,\ \hat{p}_{t+k}^h,\ \hat{\theta}_{t+k}^o,\ \hat{p}_{t+k}^o \}
\big\}.
\end{aligned}
\label{eq:goal_state}
\end{equation}
Here, $p$ denotes the position and $\theta$ indicates the rotation,  $\hat{\cdot}$ denotes kinematic reconstruction quantities and $\ominus$ indicates rotational difference.
All continuous features in $\mathbf{s}_t$ are normalized with respect to the human’s root position and facing direction.

\noindent\textbf{Action.}
The human model contains 51 actuated joints, giving an action space
\(
a_t \in \mathbb{R}^{51 \times 3}.
\)
Each action specifies a desired joint orientation in exponential-map form, which acts as a PD-control target.
The simulator converts these targets into torques applied at each joint.

\noindent\textbf{Reward.}
The reward encourages physically accurate tracking of human-object motion. At each timestep $t$, we have:
\begin{equation}
r_t =r_t^{\text{h}} *r_t^{\text{o}} *r_t^{\text{c}} *r_t^{\text{d}} *r_t^{\text{e}},
\end{equation}
where each term measures the discrepancy between predicted and reference kinematic quantities.
Specifically, the human and object terms, $r_t^{\text{h}}$ and $r_t^{\text{o}}$ respectively,  penalize pose, position, and velocity differences
$\| \theta_t^h - \hat{\theta}_t^h \|$,
$\| p_t^h - \hat{p}_t^h \|$,
$\| \theta_t^o - \hat{\theta}_t^o \|$, and
$\| p_t^o - \hat{p}_t^o \|$;
the contact term  $r_t^{\text{c}}$ aligns obtained contact states $c_t$ with reference $\hat{c}_t$ and penalizes the distance between the contact point on the object and its paired hand joint; the distance term  $r_t^{\text{d}}$ minimizes deviation in human-object proximity $\| d_t - \hat{d}_t \|$; the energy term  $r_t^{\text{e}}$ penalizes abrupt motion and sudden contact forces. Each reward is formulated as $\exp(-\lambda E)$ for each cost function $E$ with a specific hyperparameter $\lambda$. The total reward promotes motion tracking accuracy while maintaining stable and realistic interaction dynamics between the human and the object. Please see the SuppMat for more details.

\subsection{Traverse RSI and Adaptive Sampling}
\label{sec:adaptive_sampling}
The estimated HOI motions extracted from videos serve as both the initialization state and the tracking target for our physics-based HOI tracker.
However, these reconstructions are often extremely noisy; directly feeding the estimated poses to the tracker or finetuning a pretrained policy on them leads to unstable behavior and
rollout failures.
Despite this noise, we observe that the kinematic sequence still contains valuable signals from reliable frames,
\eg, when the
contact regions are less occluded, or when the interaction motion is slow.
Initializing a rollout from a highly noisy frame often results in immediate termination, for example due to missing contact or floating objects.
In contrast, starting from a cleaner frame allows the policy to reproduce physically valid interactions for longer.
Empirically, we find that the accuracy of the contact pattern for the initial frame
is strongly correlated with the resulting rollout duration in early RL training.

To detect these
reliable
kinematic frames, we adopt Reference State Initialization (RSI)~\cite{peng2018deepmimic}, which sets the simulator state $\mathbf{q}_t$ to a reference pose $\hat{\mathbf{q}}_t$ at a randomly sampled timestep.
Unlike standard RSI, which focuses on improving later-phase tracking, our goal is to identify
the potentially reliable frames for subsequent training.
Thus, during the early phase of RL, we uniformly sample initialization states across the entire time axis so every frame is visited equally often, a strategy we call \emph{Traverse RSI}.

After training for several epochs, we can observe a clear separation: rollouts initialized from reliable frames achieve substantially longer rollouts before failure, while those from noisy frames collapse almost immediately. Please see the SuppMat for the condition of rollout termination.
To identify the reliable frames,
we maintain a buffer that records the rollout length associated with the initialization from each frame.
Using the information of rollout length, we introduce an \emph{adaptive} sampling strategy that progressively increases the sampling probability of cleaner frames.
This ensures that RSI selects informative frames which facilitate stable and effective policy learning.

\begin{figure*}
\centering
\vspace{-1.5em}
\includegraphics[width=1.0\textwidth]{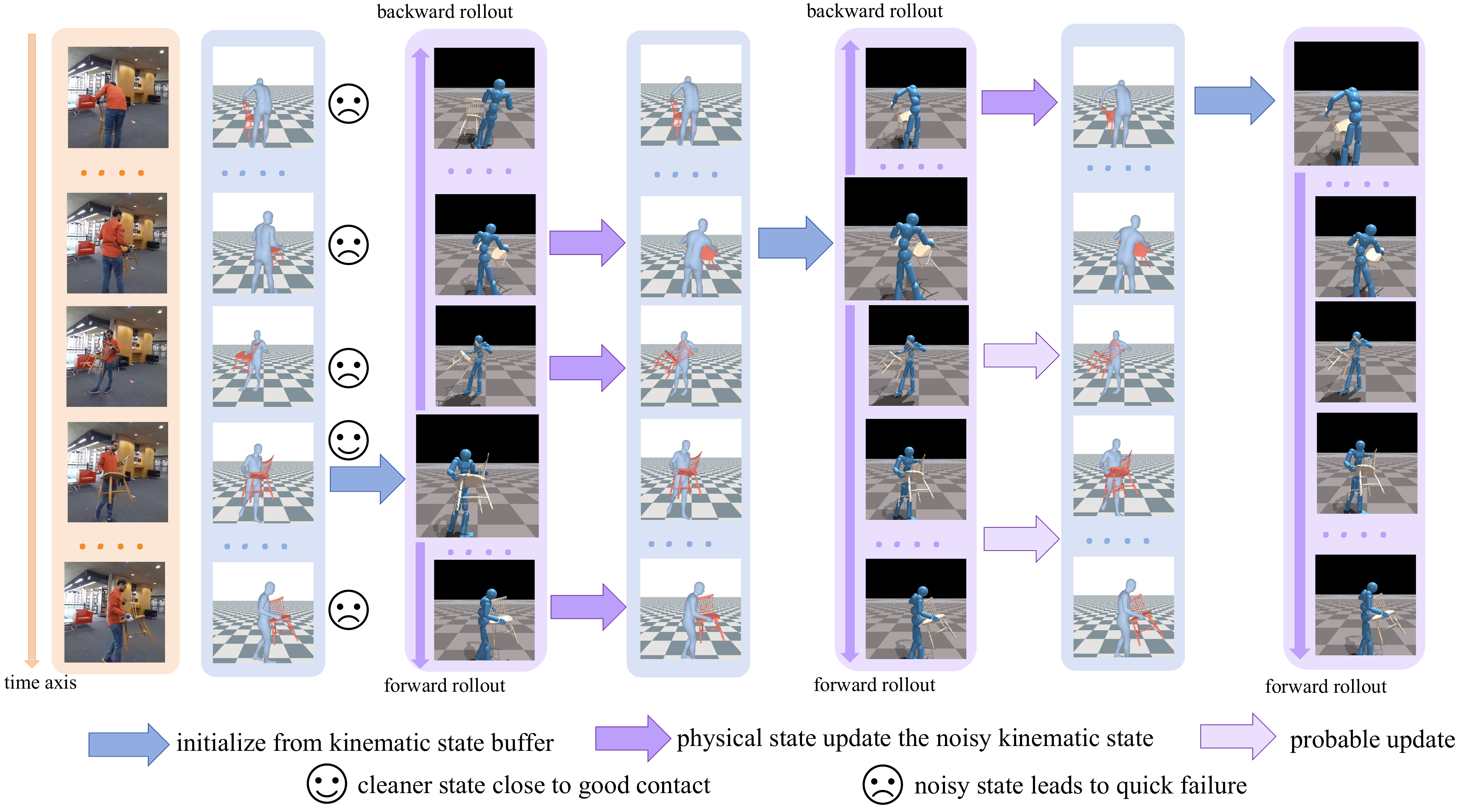}
\vspace{-1.8em}
\caption{
\textbf{Overview of our dual propagation with kinematics update mechanism.}
Kinematic estimates from monocular videos are often highly noisy.
Rollouts initialized from these noisy states typically fail quickly, whereas rollouts that start from frames with accurate contact configurations succeed for much longer.
To propagate these physically plausible states across the sequence, we train two HOI tracking policies simultaneously: a \emph{forward policy} that performs forward rollouts and a \emph{backward policy} that performs backward rollouts.
States from the successful portions of previous rollouts are used to \emph{update} the corresponding noisy kinematic frames, and subsequent rollouts initialize from these improved states.
For simplified illustration, we show three iterations of this evolving propagate-update cycle.
Overall, our dual propagation and kinematics update mechanism enables the policy to learn from extremely noisy reconstruction results and gradually recover the entire HOI sequence in a physically plausible manner.
}
\vspace{-1.5em}
\label{fig:dual_propogate}
\end{figure*}

\subsection{Dual Propagation with Kinematics Update}
After training with adaptive sampling, rollouts initialized from cleaner frames begin to succeed for more timesteps, even when the kinematic reference frames remain noisy.
As a result, the rollout often produces states that are more physically plausible compared to the original kinematic references.
We refer to this phenomenon as \textit{propagation of accurate kinematic states}.

We exploit this by maintaining a buffer that stores rollout states and updated kinematic states.
During rollouts, we record the valid simulated states up to the timestep where the rollout terminates.
When the contact configuration is accurate, these newly obtained states are often more physically plausible than the original estimates from the kinematic reconstruction (\ie, VisTracker).
These refined states are used to update the initial noisy kinematic sequence, and future rollouts sample their kinematic references from this continually evolving buffer, using a probability distribution weighted by rollout success statistics, similar to Sec.~\ref{sec:adaptive_sampling}.
Frames are updated based on the remaining rollout length and reward.
Inspired by~\cite{xu2025parc}, the state sequences obtained from previous rollouts serve not only as initialization states but also as new tracking targets.

Interestingly, this propagation occurs not only forward in time but also backward.
Starting from a physically valid contact configuration, we can propagate it to earlier frames by replaying the video in reverse.
Thus, recovering a reliable contact state at any point in the sequence enables refinement of both future and past interactions.
To leverage this insight, our implementation trains two tracking policies simultaneously: a \emph{forward policy}, which tracks the sequence chronologically, and a \emph{backward policy}, which tracks it in reverse.
As described in Section~\ref{sec:hoi_tracking}, the forward policy receives the current physics state together with future kinematic reference states, whereas the backward policy receives the current physics state and past references.
Both policies are initialized using the InterMimic~\cite{xu2025intermimic} policy checkpoint and we further finetune them.

As shown in Figure~\ref{fig:dual_propogate}, the states generated from forward and backward rollouts are used to update the initial noisy kinematic sequence.
Initializing new rollouts from these refined states improves policy learning compared to starting from the noisy vision-estimated states that lead to quick failures.
The forward policy can initialize from states refined by previous backward rollouts, and the backward policy can also initialize from states improved by previous forward rollouts.
This is particularly effective when a transition is easier to obtain in one temporal direction than the other.
For example, we observe that placing an object down is easier for the policy than picking it up from noisy kinematic references. Therefore, the backward rollout of placing the object provides a superior tracking target for the forward policy attempting the pickup.
Through this \emph{dual propagation} process, the updates gradually expand across the entire sequence until both policies can reconstruct the whole HOI sequence in a physically consistent manner.
Please refer to the SuppMat for more details of the algorithm.

%% file: sec/4_experiments.tex
\section{Experimental Evaluation}

\label{sec:experiments}
\begin{table*}[t]
\centering
\setlength{\tabcolsep}{4pt}
\renewcommand{\arraystretch}{1.2}
\small

\begin{tabular}{cccccccccc}
\toprule
 Dataset&Method & CD-H$\downarrow$& CD-O$\downarrow$& ContRate-h$\uparrow$& ContDist-h$\downarrow$& ContDist-w$\downarrow$& Pen$\downarrow$&ObjFloat$\downarrow$&ObjJerk$\downarrow$\\
 \midrule
\multirow{2}{*}{BEHAVE}&VisTracker~\cite{xie2023visibility}& \textbf{5.39}& \textbf{8.73}& 0.52& 7.78& 7.23&  6.64&0.30&524.9\\
 &Ours & 6.82& 11.06& \textbf{0.89}& \textbf{4.33}& \textbf{4.14}&  \textbf{3.91}&\textbf{0.10}&\textbf{188.5}\\
 \midrule
 \multirow{2}{*}{InterCap}&VisTracker~\cite{xie2023visibility}& \textbf{6.39}& \textbf{11.07}& 0.48& 10.22& 9.79&  3.11&0.49&508.2\\
 &Ours & 7.04& 12.32& \textbf{0.81}& \textbf{4.84}& \textbf{6.62}&  \textbf{1.76}&\textbf{0.06}&\textbf{151.2}\\
\bottomrule
\end{tabular}

\vspace{-1.0em}
\caption{
\textbf{Quantitative comparison with kinematics-based approaches on the BEHAVE and InterCap datasets.}
We compare against the state-of-the-art VisTracker~\cite{xie2023visibility}, which also serves as the initialization for our method.
While our approach introduces a slight degradation in the 3D accuracy metrics (CD), it consistently achieves substantial improvements on all interaction-related and physics-aware metrics.
All results are reported using the sequences of frames that our physical tracker successfully rolls out. Please refer to the SuppleMat for these successful frames of the sequences.
}
\vspace{-1.0em}
\label{tab:comparison-kin}
\end{table*}

\subsection{Datasets}
We conduct experiments on the BEHAVE~\cite{bhatnagar2022behave} and InterCap~\cite{huang2022intercap} datasets.
BEHAVE captures 7 subjects interacting with 20 objects in natural environments and provides SMPL and object pose annotations at 1\,fps.
Following VisTracker~\cite{xie2023visibility}, we use the extended BEHAVE dataset, which registers SMPL bodies and object templates at 30\,fps.
We evaluate our method and baselines on 35 clips of the subject 03 from the extended BEHAVE test set.
InterCap similarly captures 10 subjects interacting with 10 objects, and comes with pseudo ground truth SMPL and object annotations at 30\,fps.
We follow the same test split as VisTracker and evaluate on 38 clips from the InterCap test set.

\subsection{Metrics}
To compare our approach with previous kinematic-based methods (e.g.,~\cite{xie2023visibility}), we report a set of
raw 3D accuracy metrics and physics-aware metrics that evaluate contact, penetration, object floating, and motion smoothness.
Below we describe each metric in detail.

\noindent\textbf{Chamfer Distance.}
We compute the Chamfer distance between the reconstructed and ground-truth point clouds.
Following VisTracker~\cite{xie2023visibility}, we report Procrustes-Aligned (PA) Chamfer Distance (cm) using a sliding window of 10 frames.
We compute this separately for the SMPL-H body (CD-H) and the object mesh (CD-O), after performing Procrustes Alignment on the combined human-object mesh.

\noindent\textbf{Contact Rate.}
Contact Rate (ContRate) provides a coarse measure of contact accuracy.
For all frames where the hand is in contact with the object in the ground truth,
we check whether contact is also detected in the reconstruction.
A frame is considered to have contact if the minimum distance between any hand vertex and the closest object vertex is less than 1\,cm.
ContRate is the ratio of correctly detected contact frames over all ground-truth contact frames.

\noindent\textbf{Contact Distance.}
Contact Distance (ContDist) evaluates contact more precisely.
We identify human vertices in contact with the object in the ground-truth mesh (minimum distance $<1$\,cm).
For these vertices, we compute their unsigned distance to the closest object vertex in the reconstruction.
We report this metric at both the hand level and the whole-body level.

\noindent\textbf{Penetration.}
The penetration metric (Pen) measures human-object intersections.
For each frame, we compute the maximum penetration depth (cm) of any human vertex into the object using the object's signed distance field (SDF),
and then average across frames with penetration.

\noindent\textbf{Object Floating.}
Object Floating (ObjFloat) evaluates whether the reconstructed object violates physical support.
We consider only frames where, in the ground truth, the object is not in contact with the ground.
Among these frames, we count how many reconstructed frames show the object not being supported by the human (i.e., not in contact with the human).
ObjFloat is the ratio of unsupported frames over all frames where the object is in the air.

\noindent\textbf{Object Jerk.}
The object jerk metric (ObjJerk) assesses motion smoothness.
We compute jerk as the time derivative of acceleration for each object vertex and report the mean jerk across all vertices.

\medskip
\noindent In addition to kinematic-based methods, we also compare against physics-based approaches (e.g.,~\cite{xu2025intermimic}).
For these baselines, we adopt a subset of the 3D metrics and evaluate skill-learning performance using two success rate metrics:

\noindent\textbf{Success Rate-Binary (SR-B).}
SR-B measures whether the policy can successfully reconstruct the entire HOI sequence from start to end without failure.

\noindent\textbf{Success Rate-Frame (SR-F).}
SR-F measures the maximum continuous segment of frames that the policy can reconstruct successfully, divided by the total sequence length.
Rollouts shorter than 2 seconds are ignored.
This metric reflects the longest physically valid portion of the sequence that the policy can reproduce.

\begin{figure}
\centering
\vspace{-1.0em}
\includegraphics[width=0.45\textwidth]{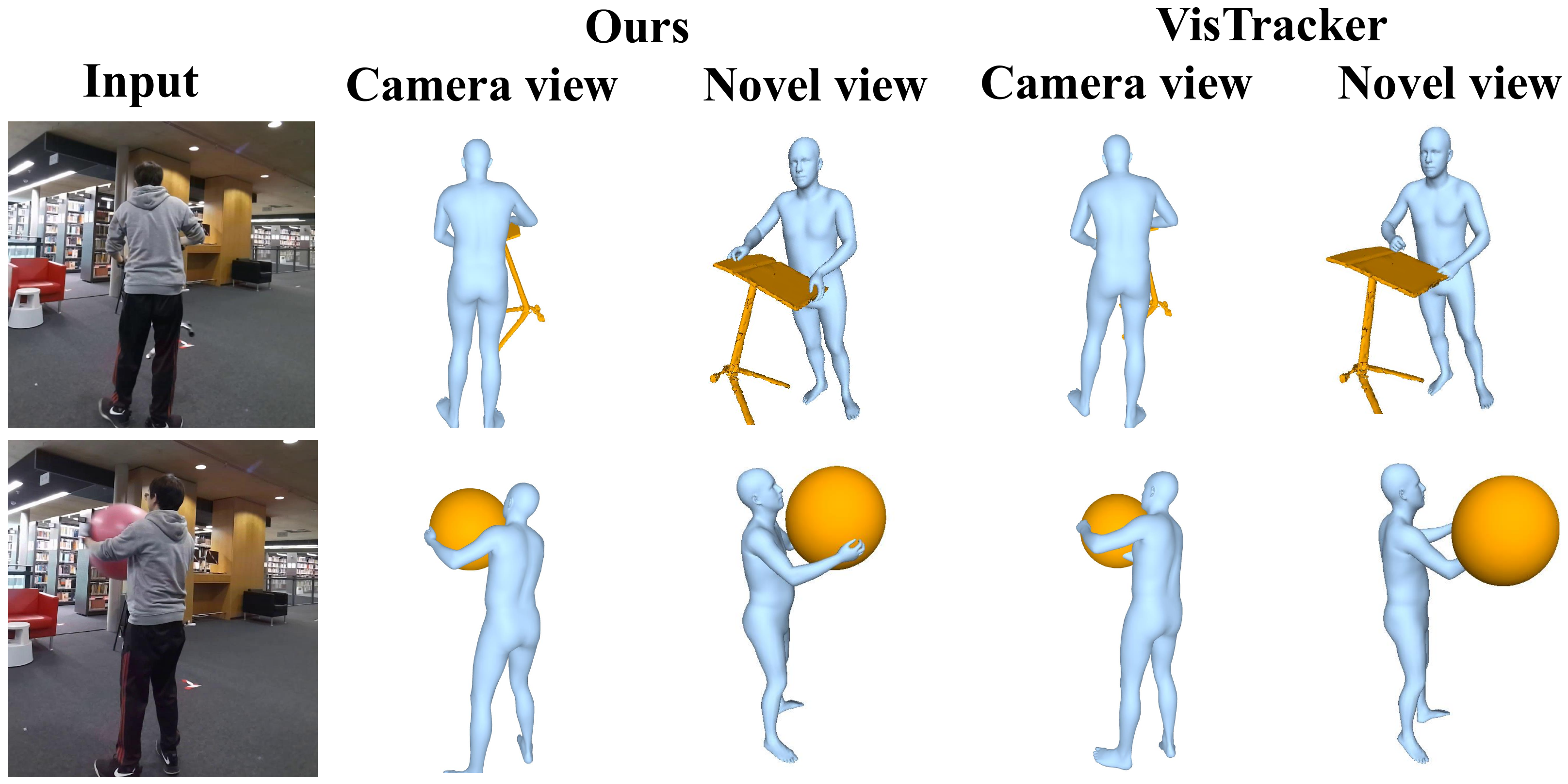}
\vspace{-1.0em}
\caption{
\textbf{Qualitative comparison with the kinematics-based method~\cite{xie2023visibility}.}
Our method successfully resolves the issues of contact floating and penetration present in the baseline, producing physically plausible human-object interaction reconstructions.
}
\vspace{-1.5em}
\label{fig:qualitative-kin}
\end{figure}

\begin{table*}[t]
\centering
\setlength{\tabcolsep}{4pt}
\renewcommand{\arraystretch}{1.2}
\small
\vspace{-1.0em}
\begin{tabular}{lcccccccc}
\toprule
 Dataset&Method & SR - B $\uparrow$&SR - F$\uparrow$& CD-H$\downarrow$& CD-O$\downarrow$& ContRate-h$\uparrow$&ContDist-h$\downarrow$   &ContDist-w$\downarrow$\\
\midrule
 &InterMimic (direct inference)&  0&3.8& -& -&  -& - &-\\
 BEHAVE& InterMimic (finetune)&  17.1&26.7& 7.10& 12.48&  0.82& 6.40&6.40\\
 &Ours &  \textbf{51.4}&\textbf{60.0}& \textbf{6.74}& \textbf{10.50}&  \textbf{0.87}& \textbf{5.99}&\textbf{5.19}\\
\midrule
 &InterMimic (direct inference)&  0&8.8& -& -&  -&  -&-\\
 InterCap&InterMimic (finetune)&  21.1&29.5& \textbf{6.32}& 12.29&  0.70&  8.46&9.79\\
 &Ours &  \textbf{52.6}&\textbf{57.1}& 6.45& \textbf{10.54}&  \textbf{0.71}&  \textbf{6.64}&\textbf{9.25}\\
\bottomrule
\end{tabular}
\vspace{-1.0em}
\caption{\textbf{Quantitative comparison with physics-based approaches on the BEHAVE and InterCap datasets.} We compare with the state-of-the-art InterMimic~\cite{xu2025intermimic} approach, when it is used for direct inference, as well as when it is finetuned.
All approaches are initialized using the VisTracker~\cite{xie2023visibility} estimate.
Given the low success rate for the direct inference version of InterMimic, we do not report 3D metrics (\ie, CD and Contact metrics) for it.
Also, these 3D metrics are evaluated on the intersection of successful frames of InterMimic (finetune) and our method.
Interestingly, our approach outperforms both baselines in the large majority of metrics. Please refer to the SuppMat for the intersection of successful frames of the sequences.
}
\label{tab:comparison-phy}
\vspace{-1.5em}
\end{table*}

\subsection{Comparison with Kinematics-based Baselines}
We use the kinematics-based method VisTracker~\cite{xie2023visibility} as our primary baseline.
VisTracker is a state-of-the-art video-based HOI reconstruction approach that relies on a provided template of the object geometry to estimate human-object interactions.
For fair comparison, we evaluate all methods using physics-aware 3D metrics on the continuous clips that the policy can successfully roll out, following~\cite{li2025maniptrans,hsieh2025dexman}.

\noindent\textbf{Quantitative Comparison.}
Table~\ref{tab:comparison-kin} presents the quantitative results.
While our method introduces a slight degradation in the Chamfer Distance metrics (approximately 1.4\,cm),
it achieves substantial improvements over VisTracker on all interaction-related and physics-aware metrics across both BEHAVE and InterCap.
These results demonstrate that our approach can recover \emph{physically plausible} and \emph{dynamically consistent} HOI sequences from monocular videos while still maintaining accurate 3D poses for the human and the object.

\noindent\textbf{Qualitative Comparison.}
In Figure~\ref{fig:qualitative-kin}, we show qualitative results comparing our method against VisTracker~\cite{xie2023visibility}.
For visualization, we transform our humanoid pose to SMPL-H pose parameters and generate the corresponding body mesh using the initial shape parameters.
VisTracker exhibits severe artifacts such as floating and penetration.
In contrast, our method produces a \emph{physically grounded} HOI sequence, even though it is initialized with noisy VisTracker reconstructions, demonstrating its ability to correct kinematic errors and enforce realistic contact dynamics.

\begin{figure*}
\centering
\includegraphics[width=1.0\textwidth]{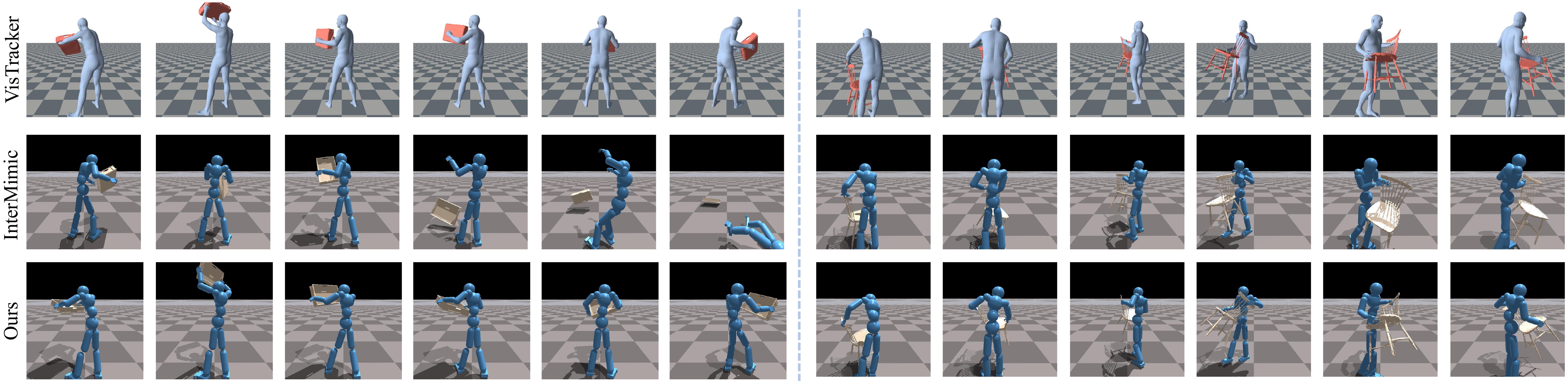}
\vspace{-2.0em}
\caption{
\textbf{Qualitative comparison with InterMimic~\cite{xu2025intermimic}.}
InterMimic often struggles to recover the correct contact configuration, due to lack of contact
in the early phase of the sequence (left) or
committing to
an unnatural contact pose that allows partial completion but does not match the interaction in the video (right).
In contrast, our method reconstructs the full sequence with physically plausible contact.
These improvements stem from our adaptive sampling and dual propagation with kinematics update, which enable the policy to overcome noisy visual inputs and maintain realistic interaction dynamics.
}
\label{fig:qualitative-phy}
\vspace{-1.0em}
\end{figure*}

\subsection{Comparison with Physics-Based Baselines}
For the comparison with physics-based methods, we adopt InterMimic~\cite{xu2025intermimic} as our baseline.
InterMimic is a state-of-the-art HOI physical tracker pretrained on the large-scale HOI motion dataset of~\cite{li2023object} and can handle interactions with diverse objects.
For fairness, we evaluate both \emph{direct inference} and \emph{single-sequence finetuning}.
All policies take as input the kinematic reconstructions estimated by VisTracker~\cite{xie2023visibility}.
We use identical simulator parameters for all methods and match the training settings between our approach and the InterMimic finetuning stage.

\noindent\textbf{Quantitative Comparison.}
Table~\ref{tab:comparison-phy} summarizes the quantitative results.
Our method consistently outperforms the physics-based baselines across almost all metrics on both BEHAVE and InterCap.
We find that direct application of the pretrained InterMimic policy is nearly infeasible on the noisy VisTracker reconstructions, typically failing immediately due to unstable or missing contact.
Although single-sequence finetuning improves its performance, it still yields relatively low success rates.
In contrast, our method achieves substantial improvements in both binary success rate (SR-B) and frame-level success rate (SR-F), demonstrating strong robustness to noisy kinematic inputs and an ability to recover physically plausible HOI sequences.
Due to the extremely low success rate of InterMimic's direct inference, we omit its 3D metrics.
For fairness, the 3D metrics for InterMimic's finetuning and our method are computed only on the intersection of successful clips (which explains the difference with the results in Table~\ref{tab:comparison-kin}).
As shown in Table~\ref{tab:comparison-phy}, our method consistently achieves superior performance.
Despite being a physics-based approach, InterMimic continues to struggle with contact-related metrics, which we analyze further in the qualitative comparison.

\noindent\textbf{Qualitative Comparison.}
In Figure~\ref{fig:qualitative-phy}, we present qualitative comparisons between our method and InterMimic~\cite{xu2025intermimic}.
In several sequences (e.g., Figure~\ref{fig:qualitative-phy}, left), InterMimic fails to recover the correct contact pose due to contact floating during the starting phase.
In other cases (e.g., Figure~\ref{fig:qualitative-phy}, right), InterMimic finds an unnatural contact configuration that permits partial completion of the sequence but does not match the physical interaction shown in the video.
In contrast, our method reconstructs the entire sequence with plausible interaction geometry.
These improvements arise from our adaptive sampling and dual propagation with kinematics update, which allow our policy to overcome noisy visual reconstructions and maintain realistic contact dynamics.

\subsection{Ablation Study}
Finally, in Table~\ref{tab:ablation}, we evaluate key design decisions of our pipeline on the BEHAVE dataset.
First, we see that naive training (first row) achieves low success rates, which clearly improve when we add adaptive sampling (second row) and when we update the kinematics during training (third row).
The full version of our algorithm with dual propagation (sixth row) achieves the best results.
To further justify our design choices, we ablate two aspects of the full method.
If we update the kinematics only for initialization, but not as a tracking target (fourth row), we observe a consistent drop in success rates.
Finally, using dual propagation with a single policy instead of two (fifth row) also has a smaller negative effect on performance.

\begin{table}[t]
\centering
\small
\setlength{\tabcolsep}{6pt}
\renewcommand{\arraystretch}{1.1}
\begin{tabular}{ccc|cc}
\toprule
\makecell{Adaptive\\Sampling} & \makecell{Kinematic\\Updates} & \makecell{Dual\\Propagation} & SR-B$\uparrow$ & SR-F$\uparrow$ \\
\midrule
\no & \no & \no & 11.4 & 24.5 \\
\yes & \no & \no & 14.3& 40.4\\
\yes & \yes & \no & 17.1& 43.5\\
\yes & \pmark & \yes & 40.0& 59.4\\
\yes & \yes & \pmark & 48.5& 59.7\\
\yes & \yes & \yes & \textbf{51.4}& \textbf{60.0}\\
\bottomrule
\end{tabular}
\vspace{-1.0em}
\caption{
\textbf{Ablation study for key design choices of our approach.} Success rates are reported on BEHAVE.
{\color{darkorange}(\ding{51})} for ``Kinematic Updates'': kinematics are updated for initialization only, but \emph{not} as tracking targets.
{\color{darkorange}(\ding{51})} for ``Dual Propagation'': dual propagation is implemented with a single policy instead of two.
}
\label{tab:ablation}
\end{table}

\begin{figure}
\centering
\vspace{-1.0em}
\includegraphics[width=0.45\textwidth,
                 trim={0 0 0 0cm}, clip]{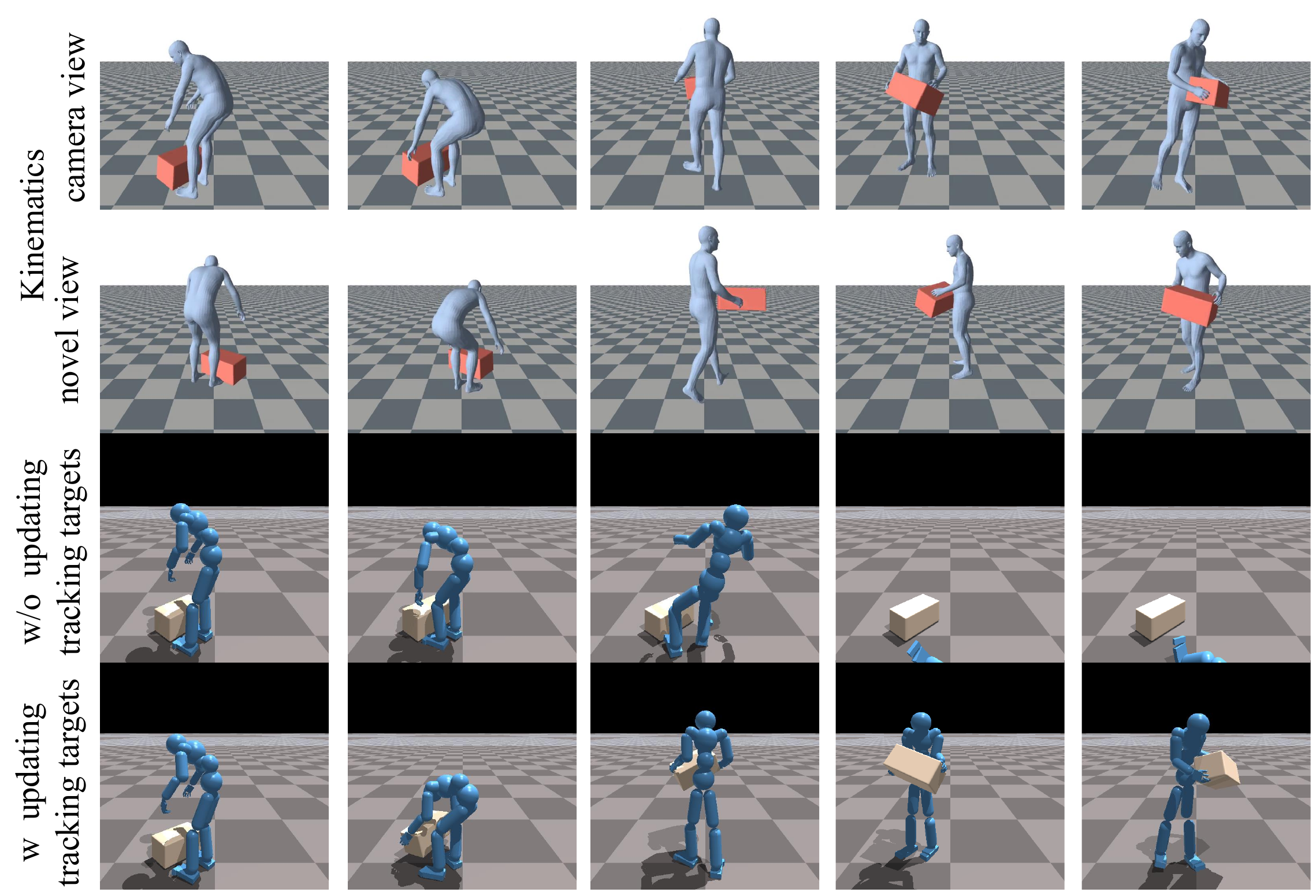}
\vspace{-1.0em}
\caption{
\textbf{Qualitative results from the ablation study.}
Without using the kinematic update as the tracking target, the policy attempts to imitate the noisy kinematic states, making it difficult to learn how to pick up the box, since there is no human-object contact during the pickup phase (second frame).
In contrast, when the kinematic update from the backward rollouts is used as the tracking target, the policy successfully learns to grasp the box and complete the subsequent actions.
}
\vspace{-1.5em}
\label{fig:ablation}
\end{figure}

%% file: sec/5_conclusion.tex
\section{Conclusion}
\label{sec:conclusion}
In this paper, we addressed the problem of recovering \emph{physically plausible, whole-body human-object interactions} from monocular videos.
We proposed a two-stage framework in which the noisy HOI motion produced by a kinematic reconstruction method is fed into a physics simulator and corrected using reinforcement learning.
The central challenge of this setting lies in the severe noise present in the initial kinematic estimates.
To address this, we introduced a novel \emph{adaptive sampling} strategy together with a \emph{dual propagation and kinematics update} mechanism that identifies reliable signals within the noisy sequence and gradually improves the entire HOI motion until it becomes physically consistent.
Through extensive experiments on benchmark datasets, we demonstrated that our method outperforms both kinematics-only and physics-based baselines across a range of physics-aware and 3D accuracy metrics.
We hope this work helps narrow the gap between vision-based HOI reconstruction and physics-based character animation, enabling more seamless real-to-sim pipelines and potentially allowing robotic systems to learn HOI skills from Internet-scale human video data.

\noindent
\textbf{Limitations and Future Work.}
A primary limitation of our approach is its two-stage pipeline, in which HOI kinematics are first reconstructed from video and then refined by a physics-based tracker.
While our kinematics update mechanism substantially improves the initial estimates, the overall success rate remains constrained by the quality of the initial 4D reconstruction.
Future work could explore developing an end-to-end video-to-dynamics system that jointly infers
geometry, and physically grounded whole-body interactions.
Another limitation of our approach is that we only handle one object per sequence with less dynamic contacts. Future work could explore more complex scenarios, including multi-object and multi-human interactions, more dynamic contacts, scene-aware HOI, etc.

\footnotesize
\noindent
\textbf{Acknowledgements:}
This project was supported by NSF-2047677, 2310666, 2413161, 2504906, 2515626, and 2544200; gifts from Adobe and Google; and computing support on the Vista GPU Cluster through the Center for Generative AI (CGAI) and the Texas Advanced Computing Center (TACC) at
the University of Texas at Austin.